\documentclass{article}

 \usepackage[preprint]{neurips_2025}

\usepackage[utf8]{inputenc} 
\usepackage[T1]{fontenc}    
\usepackage{hyperref}       
\usepackage{url}            
\usepackage{booktabs}       
\usepackage{amsfonts}       
\usepackage{nicefrac}       
\usepackage{microtype}      
\usepackage{xcolor}         

\title{\myname{}: Adaptive Rubric Learning via Memory Tuning and Adversarial Probing}

\author{
Yifei Xu\textsuperscript{1,2}, 
Guilherme Potje\textsuperscript{1}, 
Shivam Shandilya\textsuperscript{1}, 
Tiancheng Yuan\textsuperscript{1} \\
\textbf{
Leonardo de Oliveira Nunes\textsuperscript{1}, 
Rakshanda Agarwal\textsuperscript{1}, 
Saeid Asgari\textsuperscript{1}, 
Adam Atkinson\textsuperscript{1}} \\
\textbf{
Emre K{\i}c{\i}man\textsuperscript{1}, 
Songwu Lu\textsuperscript{2}, 
Ranveer Chandra\textsuperscript{1}, 
Tusher Chakraborty\textsuperscript{1}} \\
$^{1}$Microsoft \quad 
$^{2}$University of California, Los Angeles
}
\usepackage{amsmath}
\usepackage{graphicx}
\usepackage{enumitem}
\usepackage{multirow}
\usepackage{subcaption}
\usepackage[most]{tcolorbox}
\usepackage{tabularx}
\usepackage{array}
\usepackage{booktabs}
\usepackage{fvextra}
\usepackage{ltablex}

\newcommand{\bbb}[1]{\noindent\textbf{#1}}

\newcommand{\myname}[1]{$\bf{SibylSense}$#1}
\newcommand{\rar}[0]{RaR-Medicine}
\newcommand{\gov}[0]{GovReport}

\newcolumntype{Y}{>{\raggedright\arraybackslash}X}
\newcolumntype{L}[1]{>{\raggedright\arraybackslash}p{#1}}

\newcommand{\trace}[1]{{\color{black!50}\emph{Reasoning excerpt: ``#1''}}}

\begin{document}

\maketitle

\begingroup
\renewcommand\thefootnote{}%
\footnotetext{Correspondence: yxu@cs.ucla.edu, tusher.chakraborty@microsoft.com}%
\endgroup

\begin{abstract}
Designing aligned and robust rewards for open-ended generation remains a key barrier to RL post-training. Rubrics provide structured, interpretable supervision, but scaling rubric construction is difficult: expert rubrics are costly, prompted rubrics are often superficial or inconsistent, and fixed-pool discriminative rubrics can saturate and drift, enabling reward hacking. 
We present \textbf{\myname{}}, an inference-time learning approach that adapts a frozen rubric generator through a tunable memory bank of validated rubric items. Memory is updated via verifier-based item rewards measured by reference--candidate answer discriminative gaps from a handful of examples. \myname{} alternates memory tuning with a rubric-adversarial policy update that produces rubric-satisfying candidate answers, shrinking discriminative gaps and driving the rubric generator to capture new quality dimensions. 
Experiments on two open-ended tasks show that \myname{} yields more discriminative rubrics and improves downstream RL performance over static and non-adaptive baselines.
\end{abstract}

\section{Introduction}

Large language models (LLMs) can improve markedly during post-training when optimization is guided by reliable feedback signals. For example, reinforcement learning with verifiable rewards (RLVR) scales in structured domains such as mathematics and programming, where deterministic scoring rules or unit tests provide objective supervision~\cite{guo2025deepseek, shao2024deepseekmath, liu2025code}. By contrast, many high-impact applications are open-ended (e.g., document drafting and revision, analytical reporting, and question answering) where success criteria are task-dependent and inherently multi-dimensional (e.g., clarity, compliance, stylistic fidelity). Because such notions of quality are rarely captured by ground-truth similarity, designing reward functions that are simultaneously aligned and robust remains challenging~\cite{su2025crossing, xu2025direct}. Recent studies show that rubrics address this gap by decomposing quality into structured, interpretable criteria across multiple dimensions, yielding richer supervision reward signals for RL-based post-training~\cite{gunjal2025rubrics, shao2025dr}.

Despite their promise, rubric-based rewards introduce new challenges when deployed at the scale required for RL post-training. Expert-authored rubrics are often prohibitively expensive, motivating automated rubric generation with LLMs via few-shot prompting~\cite{liu2025openrubrics, he2025advancedif}. However, such automatically induced rubrics are frequently superficial and inconsistent across queries, which can cause rubric scores to saturate early or encourage over-regularization to superficial, easily-optimized rubric features during training~\cite{huang2025reinforcement, zhou2025breaking}. A further complication is that effective rubrics for RL are inherently \emph{policy-dependent}, e.g., the useful criteria often target failure modes exhibited by the current policy’s candidate answers~\cite{wu2025rlac, viswanathan2025checklists}. As the trainee policy evolves during RL training, rubrics must evolve as well. Recent approaches therefore propose generating rubrics online during training to maximize discriminativeness or critique strength across candidate answers~\cite{shao2025dr, rezaei2025online, wu2025rlac}. Yet, without grounding, highly discriminative rubrics can become misaligned with the underlying task objective, and may not even be satisfiable by a reference answer, creating opportunities for reward hacking. In aggregate, the central challenge is to generate \emph{rubrics that are consistent and fine-grained, evolve over training, and remain grounded in the task’s desired outcomes}, all at scale.

To address these challenges, we propose \textbf{\myname{}}\footnote{\textbf{SibylSense}: \emph{Sibyl} refers to an ancient prophetess associated with oracular pronouncements; we ``sense'' these oracles by extracting grounded, interpretable rubric signals for learning.}, which reframes adaptive rubric generation as a \textbf{memory tuning} problem with frozen rubric generation model. \myname{} maintains a global memory bank of rubric items that have been empirically validated, each paired with evidence of its effectiveness across multiple queries. At rubric generation time, \myname{} retrieves the most useful prior rubric practices from this bank and uses them to condition rubric synthesis, promoting cross-instance consistency and reducing the superficiality (see Case 1 in Section~\ref{sec:case_studies}). To ensure grounding, we score each proposed rubric item with an LLM-based verifier that measures its ability to discriminate the \emph{expert reference answer} from a set of candidate answers sampled from the current policy: items that fail to separate reference from candidates (or that penalize the reference) are down-weighted. The memory bank is then updated with rubric items along with their verification feedback, and we maintain efficiency and diversity by organizing items into semantically coherent categories via a rubric categorizer. Empirically, we find that this memory tuning loop converges quickly, often requiring fewer than 10 query-reference answer pairs, suggesting that \myname{} can discover a compact set of broadly effective, grounded rubric primitives with limited supervision.

While memory tuning improves consistency and grounding, a remaining challenge is that the tuned memory may cover only the quality dimensions needed to separate the reference from the \emph{current} candidate pool. Training a policy on such rubrics can over-regularize toward these initially captured dimensions, leaving other important aspects of quality underspecified. To address this and satisfy the requirement that rubrics evolve as the policy evolves, \myname{} introduces an outer \textbf{adversarial candidate refresh} loop around inner memory tuning loop. After the memory bank converges on the current candidate pool, we perform RL to train the answer-generation policy using rubric-based rewards induced by the current memory; as the policy learns to satisfy these criteria, the reference--candidate gap (and hence verifier discriminativeness) shrinks, effectively \emph{adversarially} reducing the scores of previously effective rubric items. We then refresh the candidate pool by sampling from the newly trained policy and rerun memory tuning on these harder candidates, which increasingly satisfy previously captured dimensions and thereby expose missing ones (see Case 2 in Section~\ref{sec:case_studies}). This iterative adversarial refresh creates an \emph{implicit curriculum} that continually expands rubric coverage, improves fine-grained discrimination over time, and keeps rubric generation synchronized with the changing policy.

\noindent\textbf{Summary of Contributions:} In this paper, we make the following contributions.
\begin{itemize}[leftmargin=3mm]
    \item \textbf{Inference-time learning via memory tuning.} We cast adaptive rubric generation for open-ended tasks as an inference-time learning problem: a frozen rubric generator adapts through a tunable memory bank rather than weight updates. Memory entries are optimized using verifier-derived item rewards based on each rubric item’s discriminative power over reference--candidate gaps, yielding consistent and grounded rubric primitives.
    \item \textbf{Adversarial candidate refresh.} We introduce an adversarial candidate-refresh mechanism that alternates memory tuning with a rubric-adversarial policy update. By generating best-response candidates that exploit blind spots of the current rubrics, this loop surfaces harder failure modes and expands rubric coverage beyond the initial candidate pool.
    \item \textbf{Empirical evaluation.} We evaluate \myname{} on two open-ended tasks, showing that the induced rubrics are more discriminative and that using them as rewards yields improved downstream RL performance relative to static and non-adaptive baselines.
\end{itemize}

\section{Method}
\vspace{-1mm}

\begin{figure}
    \centering
    \includegraphics[width=1\linewidth]{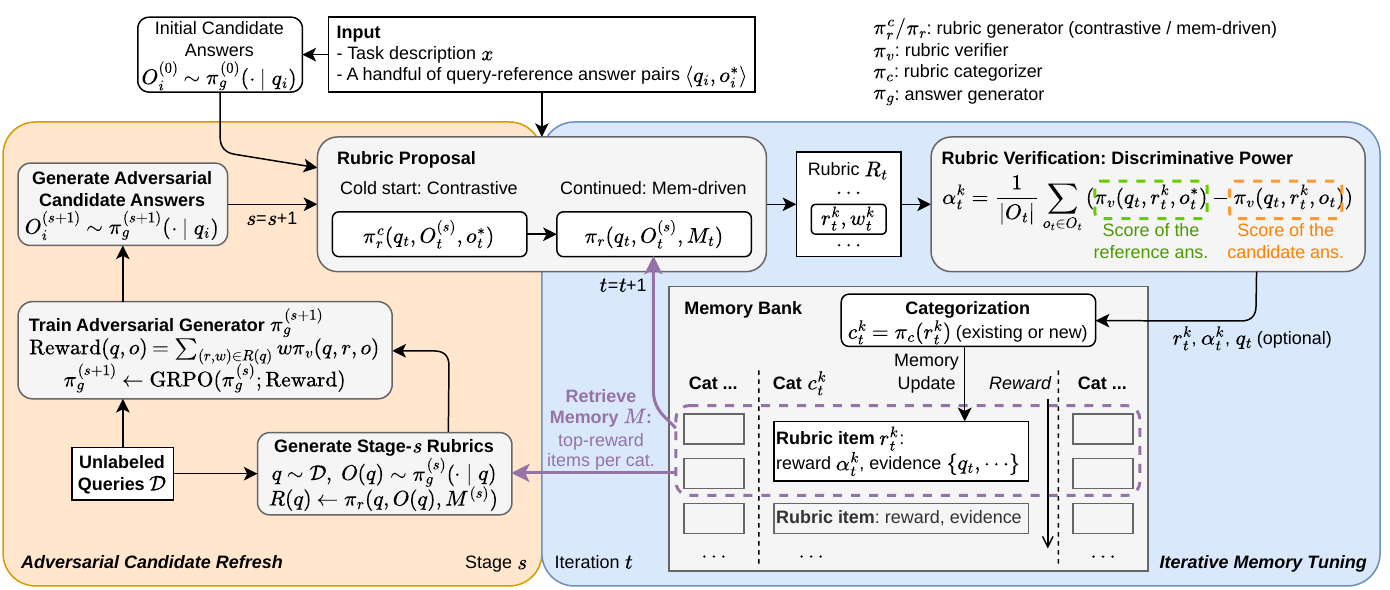}
    \caption{\textbf{\myname{}} uses frozen models and a tunable memory bank for adaptive rubric generation. The inner loop, \emph{\textcolor{blue}{iterative memory tuning}}, generates rubrics (contrastive cold start, then memory-driven), scores them with a verifier, categorizes and stores them, and retrieves top items per category to guide the next iteration. The outer loop, \emph{\textcolor{orange}{adversarial candidate refresh}}, uses the current rubrics to train an adversarial generator that produces harder candidate answers on unlabeled queries, refreshing the candidate pool and expanding failure-mode coverage.}
    \label{fig:overview}
\end{figure}

\subsection{Problem Setup: Optimizing Memory for Rubric Generation}

We study rubric generation for open-ended tasks, where evaluation criteria are underspecified and domain-dependent. Let $\mathcal{Q}$ denote a distribution of queries. Due to the scarcity of expert annotations in such domains, we assume access to a limited set of expert-verified query--answer pairs $(q, o^\star)$, which serve as reference solutions.

For each query $q \sim \mathcal{Q}$, let $\mathcal{O}(q)$ denote a distribution over candidate answers. The goal of rubric generation is to construct evaluation criteria that can reliably distinguish the reference answer $o^\star$ from other plausible candidates drawn from $\mathcal{O}(q)$.

\paragraph{Discriminative power of rubrics.}
A \emph{rubric item} $r \in \mathcal{R}$ is a natural-language statement specifying a single evaluation criterion. Each rubric item is associated with a non-negative weight $w$, and for any rubric $R$, the weights are normalized such that $\sum w = 1$. A \emph{rubric} is thus a finite set of weighted items
\begin{equation}
R = \{(r^1, w^1), \dots, (r^K, w^K)\}.
\label{eq:rubric-def}
\end{equation}
We assume access to a verifier model
\begin{equation}
\pi_v : (q, o, r) \mapsto [0,1],
\label{eq:verifier}
\end{equation}
which estimates the extent to which answer $o$ satisfies rubric item $r$.

We quantify the discriminative power of a rubric item through the expected verification gap between the reference answer and candidate answers:
\begin{equation}
\Delta(r; q)
=
\pi_v(q, o^\star, r)
-
\mathbb{E}_{o \sim \mathcal{O}(q)} \big[ \pi_v(q, o, r) \big].
\label{eq:item-gap}
\end{equation}
The discriminative power of a rubric $R$ is defined as the weighted average of item-level gaps:
\begin{equation}
\Delta(R; q)
=
\sum_{(r, w) \in R}
w \, \Delta(r; q).
\label{eq:rubric-gap}
\end{equation}
This formulation favors rubric items that are consistently satisfied by reference answers while being violated by other candidates, providing a tractable surrogate for evaluative usefulness in open-ended settings.

\paragraph{From generator optimization to memory optimization.}
Rather than directly optimizing a rubric generator, we reformulate the problem around a memory-guided generation process. Let $\mathcal{B}$ denote a global memory bank that stores structured information related to rubric construction, such as reusable evaluation practices, patterns, or heuristics. The internal representation of $\mathcal{B}$ is left unconstrained.

A retrieval function $\rho$ maps the memory bank to a natural-language memory
\begin{equation}
M = \rho(\mathcal{B}),
\label{eq:memory-retrieval}
\end{equation}
which serves as contextual guidance for rubric generation. Given a query $q$, reference answer $o^\star$, candidate answers from $\mathcal{O}(q)$, and memory $M$, a \textit{frozen rubric generator} $\pi_r$ produces a rubric:
\begin{equation}
R \sim \pi_r(\cdot \mid q, M).
\label{eq:generator}
\end{equation}
Although the memory $M$ is shared across queries, the generator conditions on $q$ and candidate answers and may utilize different aspects of $M$ depending on the instance.

We therefore shift the learning objective from optimizing the generator $\pi_r$ to optimizing the memory bank $\mathcal{B}$. Our objective is to construct a memory bank such that the rubrics generated under this memory achieve high expected discriminative power:
\begin{equation}
\max_{\mathcal{B}}
\;
\mathbb{E}_{q \sim \mathcal{Q}}
\left[
\mathbb{E}_{R \sim \pi_r(\cdot \mid q, \rho(\mathcal{B}))}
\big[
\Delta(R; q)
\big]
\right].
\label{eq:memory-objective}
\end{equation}
This objective enables optimization of rubric generation by updating the memory bank while keeping the generator and verifier fixed.

\subsection{Few-shot Iterative Memory Tuning}

We optimize the memory bank in an \emph{iterative} loop.
We start with a small set of expert-verified pairs $\{(q_i, o_i^{\star})\}_{i=1}^{I}$ (typically $I$ is small, e.g., $I=16$ in our experiments) and an initial pool of candidates for each query.
Concretely, we construct a candidate set $O_i=\{o_i^1,\dots,o_i^{J}\}$ using a base answer model $\pi_g$ (e.g., via diverse decoding):
\begin{equation}
    o_i^j \sim \pi_g(\cdot\mid q_i), \qquad j=1,\dots,J.
\label{eq:base-candidates}
\end{equation}
In this section, we treat $\{O_i\}_{i=1}^I$ as a \emph{static} set during memory tuning. These candidates may not match the idealized distribution $\mathcal{O}(q)$ in the formulation; \S\ref{sec:adversarial} later introduces mechanisms to refresh and strengthen them.

At each iteration $t$, we select a query--reference pair (e.g., by cycling through the $I$ examples) and denote it by $(q_t,o_t^{\star})$, with its associated candidate set $O_t$.
We maintain a global memory bank $\mathcal{B}_t$ and retrieve a natural-language memory
\begin{equation}
    M_t = \rho(\mathcal{B}_t),
\label{eq:iter-memory}
\end{equation}
which conditions the (frozen) rubric generators.

\subsubsection{Proposing Rubrics}
\bbb{Contrastive cold start.}
In the first pass ($t\le I$) over the few-shot examples, we allow the generator to compare candidate answers $O_t$ with the reference answer $o_t^{\star}$ and quickly elicit discriminative rubric items:
\begin{equation}
    R_t \sim \pi_r^c(\cdot\mid q_t,O_t,o_t^{\star}).
\end{equation}
This stage helps populate $\mathcal{B}_t$ with early, contrastive ``insights'' that distinguish the reference answer from the current candidates. Specifically, we update the memory bank $\mathcal{B}_t$ but do not feed the retrieved memory $\mathcal{M}_t$ to the model at this stage. This encourages $\pi_r^c$ to explore a wider range of rubric options without being guided by memory signals, thereby maximizing the diversity, and potential usefulness, of the episodic records stored in memory. We do not assume that the rubric items generated at this stage are universally optimal; rather, this phase serves to quickly warm up the memory by populating it with diverse exploratory entries, thereby providing a practical speedup for subsequent optimization.

\bbb{Memory-driven continued alignment.}
In downstream settings (e.g., test-time generation or policy optimization), reference answers are not available beyond the few expert-verified examples. Therefore, after the warm-up stage ($t>I$), we switch to generating rubrics using memory only instead of access to reference answers:
\begin{equation}
    R_t \sim \pi_r(\cdot\mid q_t,O_t,M_t).
\end{equation}
Here $\pi_r$ is instructed to compare answers in $O_t$ and propose rubric items that explain observed strengths and weaknesses, while being grounded by the retrieved memory $M_t$. This keeps memory tuning aligned with reference-less rubric generation, and allows $\mathcal{B}_t$ to accumulate aspects that may not surface during the contrastive stage.

\subsubsection{Verifying Rubrics}
For each generated rubric item $r_t^k\in R_t$, we compute a verification-based reward:
\begin{equation}
    \alpha_t^k
    =
    \frac{1}{|O_t|}\sum_{o_t\in O_t}\Big(\pi_v(q_t,o_t^{\star},r_t^k)-\pi_v(q_t,o_t,r_t^k)\Big)
    -\sigma(q_t,r_t^k).
\label{eq:item-reward}
\end{equation}
The first term is an empirical estimate of the verification gap in Eq.~\eqref{eq:item-gap}, favoring items that separate the reference answer from the current candidates; the second term is a \textit{verifiability} regularizer that favors items yielding more stable judgments under $\pi_v$.
If $\pi_v$ produces scalar scores, $\sigma(q_t,r_t^k)$ can be the standard deviation of repeated evaluations of $\pi_v(q_t,o_t,r_t^k)$. If $\pi_v$ produces binary judgments (e.g., 1 for satisfied and 0 for unsatisfied), this term can be omitted.

\subsubsection{Updating and Retrieving Memory}
\bbb{Rubric categorization.}
We organize memory $\mathcal{B}_t$ hierarchically.
At the top level, entries are grouped by rubric category (evaluation aspect), and within each category we store individual rubric items.

Each stored item keeps:
(i) the criterion itself,
(ii) an aggregated reward, and
(iii) aggregated context and supporting evidence (e.g., queries, and optionally ground truth or feedback from the verifier).
In our current implementation, the evidence is instantiated using queries only.

At iteration $t$, we process each criterion $r_t^k \in R_t$.
We use an LLM-based categorizer $\pi_c$ that is prompted with the current category list and the new criterion, and asked to either map it to an existing category or create a new one if none fits.
Let $c_t^k = \pi_c(r_t^k)$ denote the assigned category. We then update the corresponding memory entry:
\begin{equation}
\mathcal{B}_{t+1}[c_t^k][r_t^k]
\leftarrow
\mathrm{merge}\!\left(\mathcal{B}_{t}[c_t^k][r_t^k],\; \alpha_t^k,\; q_t\right),
\qquad \forall r_t^k \in R_t.
\label{eq:memory-update}
\end{equation}
If the category $c_t^k$ or criterion $r_t^k$ does not yet exist, we initialize a new entry.
If the criterion already exists in that category, we merge by updating its reward with $\alpha_t^k$ and adding the new evidence (here, $q_t$) to its aggregated evidence.

\bbb{Retrieval for the next iteration.}
To form $M_{t+1}=\rho(\mathcal{B}_{t+1})$, we perform category-aware retrieval.
Within each category, we rank rubric items by their mean reward and keep the highest-scoring ones (e.g., the top half).
We then assemble the retrieved memory by drawing from every category, together with rewards and evidence, rather than selecting only the globally highest-scoring items.

This retrieval strategy serves two goals.
First, it preserves \textit{local comparison within each evaluation aspect}, so that strong criteria are compared only against related alternatives.
Second, it maintains \textit{balance across categories}, which improves diversity and reduces the chance that minority but important aspects are consistently overlooked.
Operationally, this is our notion of ``insights'': criteria that repeatedly show stronger discriminative power and verifiability signals under $\pi_v$ on the seen examples and candidate sets, while still preserving coverage across aspects.

\subsection{Adversarial Candidate Refresh}
\label{sec:adversarial}
The reward $\alpha_t^k$ in Eq.~\eqref{eq:item-reward} is defined relative to the candidate set used during memory tuning. If this set is weak or lacks diversity, tuning may overfit to separating $o^{\star}$ from a narrow set of failure modes.

To address this, we introduce a \emph{dual-loop} procedure: an \emph{inner} iterative tuning loop (indexed by $t$ as before) with a \emph{fixed} candidate pool, and an \emph{outer} candidate-refresh loop (indexed by $s$).
Let $\{O_i^{(s)}\}_{i=1}^I$ denote the candidate pools used in outer round $s$; we treat them as static while running the inner loop and updating $\mathcal{B}$.
We monitor a held-out validation signal (e.g., the mean item reward $\alpha$ on a small validation set), and when it converges we trigger a refresh of candidates.

Concretely, let $T_s$ denote the final inner-loop iteration in outer round $s$.
After the inner loop, we obtain the final memory bank $\mathcal{B}_{T_s}^{(s)}$ and retrieve memory
$M^{(s)}=\rho(\mathcal{B}_{T_s}^{(s)})$.
For each query $q_i$, we generate a rubric
$R_i^{(s)} \gets \pi_r(q_i, O_i^{(s)}, M^{(s)})$
using the same rubric proposer as in the inner loop.
We then use this rubric to drive an adversary that generates candidates designed to challenge the current rubric:
\begin{equation}
    O_i^{(s+1)} \sim \pi_a(\cdot\mid q_i, R_i^{(s)}), \qquad i=1,\dots,I.
\label{eq:adv-candidates}
\end{equation}

Intuitively, candidate refresh is conditioned on query-specific rubrics induced by the \emph{final} memory state of the inner loop in round $s$.
This makes the refreshed candidates explicitly target the evaluation aspects currently encoded in memory, while adapting those aspects to each query.
As a result, the adversary is more likely to expose rubric-specific blind spots and produce more informative verification gaps for subsequent tuning.

We consider three practical instantiations of $\pi_a$:
(i) \textbf{prompted adversarial generation}, which conditions a base model on the rubric to elicit challenging completions;
(ii) \textbf{search-based adversarial generation} (e.g., MCTS), which explores candidate refinements guided by rubric-based signals; and
(iii) \textbf{training-time adversarial regeneration}, where an RL-tuned generator is periodically updated and used to resample hard candidates.
In all cases, the role of adversarial candidates is to expand coverage of failure modes encountered during memory tuning, rather than to approximate an exact target distribution. In our experiments, we instantiate (iii).
\section{Results}
\label{sec:results}

We evaluate whether memory-tuned rubrics improve (i) rubric-based preference decisions and (ii) downstream policy learning when used as training-time supervision.

\subsection{Experimental Setup}

\bbb{Datasets.}
We evaluate open-ended generation on two datasets: \textbf{RaR-Medicine}~\cite{gunjal2025rubrics}, a medical reasoning QA dataset of 20K prompts paired with reference answers and instance-specific rubric annotations, and \textbf{GovReport}~\cite{huang2021efficient}, a long-document summarization dataset of U.S. government reports paired with human-written executive summaries. We follow the original splits from each dataset for training and evaluation.

\bbb{Rubric generation and verification.}
We run iterative memory tuning to obtain the final memory bank $\mathcal{B}$ and the retrieved memory $M=\rho(\mathcal{B})$.
At evaluation time, for each query we generate a rubric $R$ with the frozen generator $\pi_r(\cdot\mid q, M)$, without access to the expert reference answer.
We treat the expert reference as $o^{\star}$ and sample candidate answers from a base answer model $\pi_g$.
Unless otherwise stated, memory tuning uses $I=8$ expert-verified examples and $J=4$ candidates per query.
We instantiate $\pi_r$ with Qwen3-32B, $\pi_g$ with Qwen3-8B, and the verifier/judge $\pi_v$ with GPT-4o.

\bbb{RL from rubric-based rewards.}
We train an answer policy $\pi_{\theta}$, initialized from Qwen3-8B, using GRPO~\cite{shao2024deepseekmath}.
For each training query $q$, we sample a rubric $R\sim\pi_r(\cdot\mid q, M)$ and use the verifier-weighted rubric score as the reward:
\begin{equation}
S(q,o;R)=\sum_{(r,w)\in R} w\pi_v(q,o,r).
\label{eq:rubric-score-results}
\end{equation}
We use batch size 16, 16 rollouts per prompt ($T=1.0$, $p=0.95$), learning rate $5\times 10^{-7}$ with 0.2 warmup, clipping $\epsilon=0.2$, and entropy regularization $\beta=0.001$.
Training uses TRL/DeepSpeed with vLLM rollouts on 3$\times$8 A100 GPUs.

\bbb{Metrics.}
We report two metrics.
(1) \textbf{Preference accuracy.}
For each query $q$, we compare the verifier scores of the expert reference answer $o^\star$ and a candidate answer $o$ under each rubric item $r$.
For item $r$, we assign a pairwise outcome of 1 if $\pi_v(q,o^\star,r) > \pi_v(q,o,r)$, 0.5 if they are equal, and 0 otherwise.
Preference accuracy is the weighted average of these pairwise outcomes over rubric items, using the rubric-item weights.
(2) \textbf{Pairwise win rate (vs.\ base).}
We sample answers from the GRPO-trained policy and the base model, and report the fraction of wins for the RL policy under an external judge that uses the reference answer as the gold standard. We use OpenAI o4-mini as the judge model; prompts are provided in Appendix~\ref{app:rar_judge}.

\bbb{Baselines.}
We compare against three rubric sources:
(1) \textbf{original rubrics}, i.e., the original query-specific rubrics provided with the \rar{} dataset (not available for \gov{}); and
(2) \textbf{few-shot rubrics}, produced from a prompt containing randomly sampled contrastive examples (reference + candidate answers), without memory tuning. For \rar{} we use 5 pairs, and for \gov{} we use 3 shots due to input length and context limits. The few-shot prompts are provided in Appendix~\ref{app:fewshot_prompts}.

\subsection{Main Results}

\label{sec:main-results}

\begin{figure*}[t]
    \centering
    \begin{subfigure}[t]{0.48\textwidth}
        \centering
        \includegraphics[width=\linewidth]{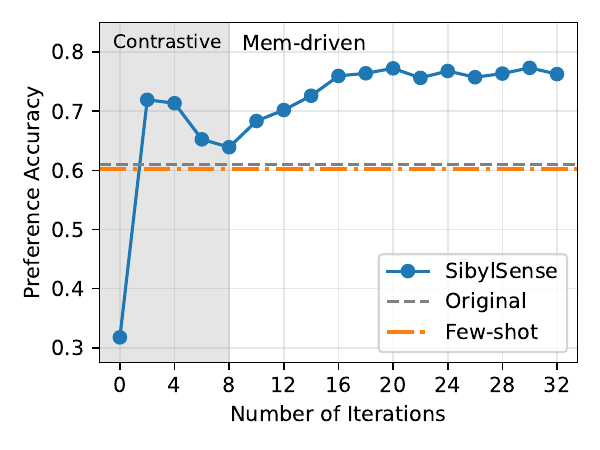}
        \caption{\rar{}.}
        \label{fig:pref-acc-rar}
    \end{subfigure}
    \hfill
    \begin{subfigure}[t]{0.48\textwidth}
        \centering
        \includegraphics[width=\linewidth]{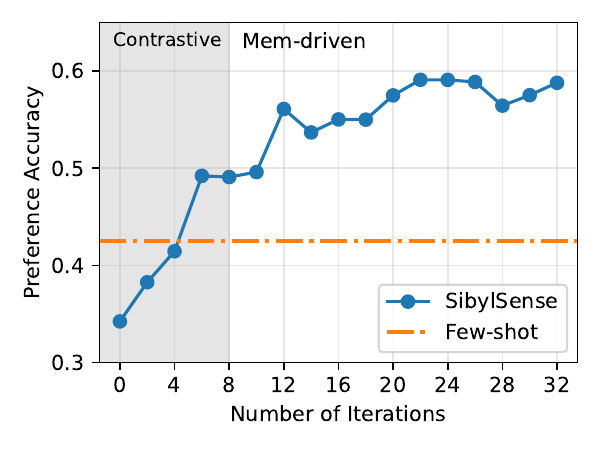}
        \caption{\gov{}.}
        \label{fig:pref-acc-gov}
    \end{subfigure}
    \caption{
    Preference accuracy vs.\ number of memory-tuning iterations on two datasets. We switch from contrastive warm-up to memory-driven (reference-free) tuning after iteration 8. Across both datasets, \myname{} becomes increasingly discriminative over iterations; contrastive warm-up accelerates early progress, while memory-driven tuning sustains improvements.}
    \label{fig:pref-acc-main}
\end{figure*}

Figure~\ref{fig:pref-acc-main} summarizes preference accuracy over memory-tuning iterations on \rar{} and \gov{}, and Table~\ref{tab:downstream-winrate} reports downstream pairwise win rate (GRPO policy vs.\ base) under an external judge.

\bbb{\myname{} consistently improves discriminative power and surpasses static rubric baselines.}
Across both datasets, preference accuracy increases with memory-tuning iterations (Fig.~\ref{fig:pref-acc-main}), indicating that the retrieved memory progressively helps the frozen rubric generator produce criteria that better separate ground-truth references from plausible candidates. In both datasets, \myname{} outperforms rubrics generated from few-shot prompting alone. On \rar{} (Fig.~\ref{fig:pref-acc-rar}), it also surpasses the original dataset-provided rubrics. On \gov{} (Fig.~\ref{fig:pref-acc-gov}), the few-shot baseline struggles to reliably prefer the ground-truth summary over model generations, likely because the desired quality dimensions are harder to identify in long-form summarization, whereas \myname{} substantially closes this gap through iterative memory tuning.

\bbb{Contrastive warm-up yields fast early gains, but memory-driven tuning stabilizes and sustains improvement.}
Because we switch from contrastive mode to memory-driven mode after iteration 8, the preference-accuracy curves exhibit a clear two-phase pattern (Fig.~\ref{fig:pref-acc-main}). The early gains are steep, consistent with the contrastive cold-start stage: direct access to reference-candidate comparisons quickly surfaces high-utility discriminative criteria. However, contrastive signals alone are not sufficient for continued improvement. On \rar{}, the contrastive-phase gains are sharp but unstable, while on \gov{} preference accuracy remains below 50\% during the contrastive phase, which is likely too weak to support reliable rubric-driven optimization.

The key limitation is that contrastive criteria are often highly case-specific and depend on access to the reference answer. At test time, however, the rubric generator must operate without references and rely on more generalizable signals. The memory-driven (reference-free) phase addresses this by accumulating and retrieving reusable, verifier-validated insights, and performance continues to improve after the warm-up switch (Fig.~\ref{fig:pref-acc-main}). Together, the curves support the view that contrastive prompting is an effective initializer, whereas memory tuning provides the stable and sustained gains needed for robust reference-free rubric generation.

\begin{table}[t]
    \centering
    \small
    \begin{tabular}{lcc}
        \toprule
        \textbf{Rubric source} & \textbf{\rar{}} & \textbf{\gov{}} \\
        \midrule
        Original Rubrics      & 49.6 & --- \\
        Few-shot Rubrics      & 51.2 & 48.9 \\
        \midrule
        \myname{}-Base        & 56.0 & 52.6 \\
        \myname{}-Adv         & \textbf{60.6} & \textbf{52.9} \\
        \bottomrule
    \end{tabular}
    \vspace{1mm}
    \caption{
    Pairwise win rate (\%, higher is better) of the GRPO-trained policy against the base model under an external judge. Rewards derived from \myname{}-generated rubrics improve downstream policy learning over static rubric baselines, and adversarial candidate refresh (\myname{}-Adv) provides an additional gain over non-adversarial memory tuning (\myname{}-Base).}
    \label{tab:downstream-winrate}

\end{table}

\bbb{Better rubrics translate to better downstream RL, and adversarial candidate refresh provides additional gains.}
The downstream win-rate results in Table~\ref{tab:downstream-winrate} show that improvements in rubric quality are not merely evaluative: they also provide stronger reward signals for policy optimization. On both datasets, GRPO policies trained with \myname{}-generated rubrics outperform policies trained with baseline rubrics, indicating that the gains in discriminative power transfer to downstream learning.

Adversarial candidate refresh further improves performance over non-adversarial memory tuning (Table~\ref{tab:downstream-winrate}), supporting the hypothesis that refreshing the candidate pool helps expose rubric-specific blind spots that are not covered by the initial samples. This effect is especially pronounced on \rar{}, while on \gov{} the gain is smaller but still positive, suggesting diminishing returns when candidate diversity is already high or the quality dimensions are more diffuse.

\subsection{Case Studies}
\label{sec:case_studies}
We present two case studies to illustrate the mechanisms behind the gains in Section~\ref{sec:main-results}: Case~1 shows criterion abstraction during memory-driven tuning, and Case~2 shows failure-mode coverage expansion from adversarial candidate refresh. For brevity, we show shortened examples and abridged reasoning excerpts.

\bbb{Case 1: memory-driven rubric proposal abstracts a narrow numerical-detail heuristic into a reusable criterion.}
This \gov{} example illustrates the mechanism behind the gains from the memory-driven phase of \myname{}. After the switch from contrastive to memory-driven proposal at iteration 8, the memory evolves from a narrow, low-reward heuristic about \emph{specific funding figures} into a broader, query-agnostic criterion about \emph{query-specific numerical data or examples not stated in the source report}. The later criterion receives a substantially higher item reward and is subsequently reused during rubric generation for a new query.

This memory evolution also has a direct impact on test-time rubric generation. Using the memory checkpoint before the memory-driven phase starts (iter-8), the proposer emits a narrower, trade-statistics-specific criterion that achieves zero item reward on the focal query. Using the memory checkpoint after memory-driven tuning (iter-32), the proposer adopts the generalized memory entry, which yields substantially higher preference accuracy (75\%). The poor candidate summary is fluent but packed with specific numeric claims (e.g., ``11\%'', ``52\%'', ``17\%'', ``9\%''), while the reference summary emphasizes higher-level framing (supply-chain dependency, congressional actions, and policy considerations), making this criterion discriminative.

\begin{tcolorbox}[
    title={\bbb{Case study 1 (GovReport): memory evolution and memory-to-rubric transfer}},
    colback=gray!3,
    colframe=black!30,
    boxrule=0.5pt,
    arc=2pt,
    left=6pt,right=6pt,top=4pt,bottom=4pt,
    breakable
]
\footnotesize
\setlength{\tabcolsep}{4pt}
\renewcommand{\arraystretch}{1.12}

\textbf{A. Memory evolution.}

\begin{tabularx}{\linewidth}{@{}L{0.17\linewidth}Y L{0.12\linewidth}@{}}
\toprule
\textbf{Stage} & \textbf{Memory entry (same category)} & \textbf{Reward} \\
\midrule
Iter-8 (end of contrastive)
& ``Avoids excessive detail on specific funding figures for individual components.'' \textit{(narrow)}
& \(+0.167\) \\

Iter-20 (memory-driven)
& ``Avoids introducing query-specific numerical data or examples not directly stated in the original report.'' \textit{(generalized)}
\par\vspace{0.25em}
\noindent\trace{...some responses mention specific years or figures, which are query-specific and should be abstracted. The rubric needs to be query-agnostic. So, instead of `mentions specific funding figures', we can say `avoids introducing query-specific numerical data'...}
& \(\mathbf{+0.500}\) \\
\bottomrule
\end{tabularx}

\vspace{0.7em}
\textbf{B. Memory-to-rubric transfer on a downstream query.}

\begin{tabularx}{\linewidth}{@{}L{0.17\linewidth}Y L{0.12\linewidth}@{}}
\toprule
\textbf{Memory checkpoint} & \textbf{Proposed rubric item} & \textbf{Preference Accuracy} \\
\midrule
Iter-8
& ``Avoids including procedural or numerical trade statistics (e.g., specific import/export values, tariff percentages) that do not directly support core findings.'' \textit{(narrow / misaligned)}
& \(50.0\%\) \\

Iter-32
& ``Avoids introducing query-specific numerical data or examples not directly stated in the original report.'' \textit{(verbatim reuse from memory)}
\par\vspace{0.25em}
\noindent\trace{...The rubric should be query-agnostic, so it doesn't rely on specific details from this report... Looking at the prior cases... low-alignment items were too specific or not discriminative... Candidate responses exhibit common failure modes such as including query-specific numerical data (e.g., `17\% drop in Chinese exports')...}
& \(\mathbf{75.0\%}\) \\
\bottomrule
\end{tabularx}

\vspace{0.7em}
\textbf{C. Reference vs. candidate (shortened).}

\begin{tabularx}{\linewidth}{@{}Y Y@{}}
\toprule
\textbf{Reference summary (high-level framing)} & \textbf{Candidate summary (numeric/detail drift)} \\
\midrule
The reference frames the report at a high level: U.S.\ dependence on China-based supply chains, medical supply shortages, congressional responses (including CARES Act provisions), and broader policy considerations such as diversification, data gaps, and U.S.\ leadership on global health/trade issues. It emphasizes report-level findings and implications rather than enumerating many instance-specific figures.
&
The candidate is fluent but introduces many specific numerical claims and examples (e.g., ``11\% of U.S.\ trade,'' ``52\% of U.S.\ imports of pharmaceutical ingredients,'' ``17\%'' export decline, ``9\%'' GDP drop), along with additional detailed examples. These specifics can distract from the report's central findings and recommendations, matching the failure mode captured by the learned memory criterion. \\
\bottomrule
\end{tabularx}

\vspace{0.5em}
\textbf{Interpretation.} The evidence shows both \emph{memory evolution} (from a narrow to a generalized criterion with higher reward) and \emph{memory-to-rubric transfer} (the generalized memory entry is later reused and becomes discriminative on a new query).
\end{tcolorbox}

\bbb{Case 2: adversarial candidate refresh adds a missing evaluative dimension for harder negatives.}
This \rar{} example illustrates how adversarial candidate refresh expands failure-mode coverage during memory tuning. Before refresh, the memory checkpoint is dominated by highly rewarded generic criteria (e.g., format adherence, direct answer presence, and avoidance of extraneous details). This profile is often sufficient for easier negatives, but it under-represents a key dimension for clinical treatment-selection cases: explicitly contrasting the recommended treatment with relevant alternatives under the case constraints.

After adversarial refresh, \myname{} adds a new memory category, \emph{Justified Treatment Comparison}, with the rubric item \emph{``Clearly contrasts the recommended treatment with alternative options and explains why it is more suitable.''} (reward \(+0.583\)). This new criterion is induced by a harder candidate that is clinically detailed and plausible, yet does not foreground the core comparative justification (mechanical thrombectomy vs.\ IV tPA given a 5-hour onset and large vessel occlusion) as clearly as the reference. 

\begin{tcolorbox}[
    title={\bbb{Case study 2 (RaR-Medicine): adversarial candidate refresh}},
    colback=gray!3,
    colframe=black!30,
    boxrule=0.5pt,
    arc=2pt,
    left=6pt,right=6pt,top=4pt,bottom=4pt,
    breakable
]
\footnotesize
\setlength{\tabcolsep}{4pt}
\renewcommand{\arraystretch}{1.12}

\textbf{A. Pre-adv memory checkpoint.}

\begin{tabularx}{\linewidth}{@{}L{0.24\linewidth}Y L{0.11\linewidth}@{}}
\toprule
\textbf{Pattern in pre-adv memory} & \textbf{Representative entries} & \textbf{Reward} \\
\midrule
Formatting / presentation
& ``Avoids markdown formatting \ldots and presents the answer in a single paragraph.''; ``Uses plain text formatting without LaTeX \ldots''; ``Limits the response to 3 concise paragraphs or fewer.''
& up to \(+1.000\) \\

Avoiding extraneous details
& ``Avoids excessive detail on adjunctive therapies or extended criteria not explicitly requested.''; ``Limits explanations to directly relevant clinical criteria \ldots without extraneous details.''
& up to \(+1.000\) \\

Direct answer placement
& ``States the final answer unambiguously in the first sentence or paragraph.''
& \(+0.280 \sim +0.345\) \\

\textbf{Missing dimension}
& No criterion explicitly rewards \textbf{comparing the recommended treatment against alternatives and justifying superiority under the case constraints} (e.g., thrombectomy vs.\ IV tPA at 5 hours with LVO).
& --- \\
\bottomrule
\end{tabularx}

\vspace{0.7em}
\textbf{B. Example query, reference, and candidate responses.}

\begin{tabularx}{\linewidth}{@{}L{0.19\linewidth}Y@{}}
\toprule
\textbf{Field} & \textbf{Content (shortened)} \\
\midrule
Query
& 69-year-old man with acute aphasia/right-sided weakness for 5 hours; imaging suggests a large vessel occlusion. What is the appropriate treatment? \\

Reference answer
& Recommends \textbf{mechanical thrombectomy}; explicitly ties the decision to the \textbf{5-hour} onset and \textbf{large vessel occlusion}, and contrasts it with \textbf{IV tPA}, which is generally most effective within \(\sim 4.5\) hours and is not suitable here. \\

Pre-adv candidate answer
& Rejected before adversarial refresh; easier to separate using generic criteria such as direct answer presence and concision / extraneous-detail control. \\

Post-adv candidate answer
& More clinically detailed and plausible: includes extended-window discussion (up to 24 hours), imaging criteria, contraindications, and adjunctive therapies. However, it dilutes the prompt-specific decision rationale and does not foreground the comparative justification (why thrombectomy is more suitable than alternatives under the stated constraints) as clearly as the reference. \\
\bottomrule
\end{tabularx}

\vspace{0.7em}
\textbf{C. New memory category after adversarial candidate refresh.}

\begin{tabularx}{\linewidth}{@{}L{0.22\linewidth}Y L{0.11\linewidth}@{}}
\toprule
\textbf{Category} & \textbf{Memory entry} & \textbf{Reward} \\
\midrule
Justified treatment comparison
& ``Clearly contrasts the recommended treatment with alternative options and explains why it is more suitable.''
\par\vspace{0.2em}
\noindent\trace{...The chosen response (reference) recommends mechanical thrombectomy and explicitly contrasts it with IV tPA (less suitable beyond 4.5 hours). The rejected responses are more detailed, but they do not foreground the treatment comparison as clearly under the case constraints...}
& \(\mathbf{+0.583}\) \\
\bottomrule
\end{tabularx}

\vspace{0.5em}
\textbf{Interpretation.} Adversarial refresh produces a harder rejected candidate that is not well separated by the pre-adv memory's dominant generic criteria. This exposes a missing evaluative dimension, \emph{comparative treatment justification}, and leads to a new high-scoring memory category, illustrating how adversarial refresh expands failure-mode coverage beyond what base-candidate tuning learns.
\end{tcolorbox}
\section{Related Work}

\bbb{Learning to generate rubrics.}
Rubrics have emerged as a practical way to extend verification signals to open-ended, less-verifiable tasks~\cite{ gunjal2025rubrics, huang2025reinforcement}.  Although expert-annotated rubrics are generally reliable~\cite{he2025advancedif}, scaling their creation demands significant human effort~\cite{liu2025openrubrics, he2025advancedif,xie2025auto}. Moreover, such rubrics are inherently incomplete: as training progresses, new failure modes can emerge~\cite{shao2025dr, rezaei2025online}, limiting their effectiveness. To mitigate these limitations, recent work leverages LLMs to automatically generate rubrics and assign corresponding weights~\cite{rezaei2025online, viswanathan2025checklists}. The rewards can then be obtained via the LLM judge score of the generated rubric items.
Auto-Rubric~\cite{xie2025auto} studies rubric construction without weight updates, refining a pool of candidate criteria and then compressing it into a compact core rubric by maximizing an information-theoretic coding-rate objective; the resulting core set is meant to generalize, but it is essentially static once distilled rather than adaptively regenerated per query. 
A separate cluster makes rubrics evolve during RL itself: 
OnlineRubrics~\cite{rezaei2025online} elicits new criteria online from pairwise comparisons between current and reference policies and uses them to update model weights, 
while RLER~\cite{shao2025dr} maintains rubrics that co-evolve with the policy during training, updating both rubrics and the policy via weight updates. 
RLAC~\cite{wu2025rlac} studies a different perspective: it targets the verification-cost bottleneck by training a critic to propose likely failure modes that an external validator checks, jointly updating generator and critic. 
We instead frame adaptive rubric generation as a memory-tuning problem without weight updates, and address practical data scarcity by adversarially generating challenging candidates to broaden the initial candidate distribution.


\bbb{Inference-time learning via agent memory}
A complementary line of work studies how LLM-based systems can adapt across attempts without updating base model weights by writing experience into a persistent memory and retrieving it later. One thread treats memory updates as a training-free learning mechanism driven by feedback. Reflexion~\cite{shinn2023reflexion} frames this as verbal reinforcement, converting task feedback into reflective text that is stored and reused to condition subsequent trials. Memento~\cite{zhou2025memento} pushes this toward a continual-learning formulation for agents via memory-based online reinforcement learning over a memory-augmented decision process, where improvement arises from read and write operations over stored experience rather than gradient updates to the underlying LLM. In parallel, other work focuses on memory substrates and management policies that make long-term retrieval effective. MemoryBank~\cite{zhong2024memorybank} studies long-term personalization by maintaining and updating user-specific memories and retrieving relevant memories in later conversations. MemTree~\cite{rezazadeh2024isolated} proposes dynamic tree-structured representations for online conversation tracking and retrieval. A-MEM~\cite{xu2025mem} develops a Zettelkasten-inspired agentic memory system that creates structured notes and dynamically indexes, links, and updates them as new information arrives. While prior work primarily uses agent memory to adapt task policies and responses, we reformulate rubric optimization as a \emph{memory tuning} problem. In particular, we leverage discriminative power as a distinctive verification signal for optimizing rubric generation, and propose rubric-specific memory organization.

\section{Conclusion}
Designing reward signals that are both aligned and robust remains a central obstacle for RL post-training in open-ended generation. Rubrics offer an appealing alternative to brittle scalar rewards by providing structured, interpretable criteria, but prior approaches struggle to construct rubrics that are consistent, fine-grained, grounded, and adaptive at scale. In this work, we introduced \textbf{\myname{}}, which reframes adaptive rubric generation as \emph{inference-time learning} with frozen models, where adaptation occurs through a tunable memory bank of empirically validated rubric items updated using verifier-based discriminative feedback. To prevent saturation and distributional overfitting to a fixed candidate pool, we further proposed an outer \emph{adversarial} candidate-evolution loop: as the policy learns to satisfy current rubric criteria, discriminative gaps shrink, forcing the rubric generator to discover new quality dimensions and failure modes. Across two open-ended tasks, our results show that \myname{} produces more discriminative rubrics and yields stronger downstream RL performance than static and non-adaptive baselines. Together, these findings suggest a practical path toward scalable, grounded reward construction for open-ended RL post-training by coupling memory-based inference-time adaptation with adversarially evolving candidates.

\bibliography{references}
\bibliographystyle{plain}

\appendix
\section{Prompts for Pairwise Win-Rate Evaluation}
\label{app:rar_judge}
\begin{tcolorbox}[
  title={\rar{}},
  colback=gray!3,
  colframe=black!30,
  boxrule=0.5pt,
  arc=2pt,
  breakable
]
\begin{Verbatim}[breaklines=true, breakanywhere=true, fontsize=\footnotesize, breaksymbolleft={}]
# Task
You are an expert evaluator for medical question answering. You will be provided with:
- **Query:** a medical QA prompt
- **Reference Answer:** an expert-written answer
- **Candidate Answers:** two candidate answers to evaluate

Your goal is to determine which candidate better aligns with the reference answer.

# Query
{query}

# Reference Answer
{reference_answer}

# Candidate Answers
- Answer A: {answer_a}
- Answer B: {answer_b}

# Instructions
1. **Understand the query:** Read the *Query* to understand the medical context, what is being asked, and any constraints.

2. **Identify the desired qualities:** Read the *Reference Answer*. Treat it as the gold standard and identify the desired qualities of a good answer.

3. **Evaluate each candidate:** Carefully examine *Answer A* and *Answer B* against the desired qualities evidenced by the reference answer. Decide which candidate better satisfies those qualities. Do **not** reward additional detail or breadth that is not directly reflected in the reference answer.

4. **Output Format:** Reason step-by-step, and respond with **only** a JSON object containing your step-by-step reasoning and final judgment in the format below:
```json
{{
  "reasoning": "Your step-by-step reasoning.",
  "judgment": "'Answer A' or 'Answer B'. If both are similarly good, choose one based on your best judgment."
}}
```
\end{Verbatim}
\end{tcolorbox}
\begin{tcolorbox}[
  title={\gov{}},
  colback=gray!3,
  colframe=black!30,
  boxrule=0.5pt,
  arc=2pt,
  breakable
]
\begin{Verbatim}[breaklines=true, breakanywhere=true, fontsize=\footnotesize, breaksymbolleft={}]
# Task
You are an expert evaluator for government report summarization. You will be provided with:
- **Report:** a government report
- **Reference Summary:** an expert-written summary
- **Candidate Summaries:** two candidate summaries to evaluate

Your goal is to determine which candidate better aligns with the reference summary.

# Report
{report}

# Reference Summary
{reference_summary}

# Candidate Summaries
- Summary A: {summary_a}
- Summary B: {summary_b}

# Instructions
1. **Understand the report:** Read the *Report* to understand its main purpose, scope, and key findings.

2. **Understand the desired qualities:** Read the *Reference Summary*. Use it as the gold standard to understand the desired qualities of a good summary (e.g., format, focus, tone, brevity, etc.).

3. **Evaluate each candidate:** Carefully examine *Summary A* and *Summary B* against the desired qualities exemplified by the reference summary. Determine which candidate better aligns with the reference summary by comparing their adherence to these qualities.
  - Penalize **style mismatches** (including tone, formatting/structure, organization, verbosity, etc.) relative to the reference.
  - Penalize **additional content** not present in the reference summary (even if they appear in the report).
  - Penalize **missing or distorted** points that *are* present in the reference summary.

4. **Tie-breaker:** If both candidates are similarly aligned with the reference summary, prefer the more concise summary.

5. **Output Format:** Reason step-by-step, and respond with **only** a JSON object containing your step-by-step reasoning and final judgment in the format below:
```json
{{
  "reasoning": "Your step-by-step reasoning.",
  "judgment": "'Summary A' or 'Summary B'."
}}
```
\end{Verbatim}
\end{tcolorbox}

\section{Prompts for the Few-shot Baseline}
\label{app:fewshot_prompts}
\begin{tcolorbox}[
  title={\rar{}},
  colback=gray!3,
  colframe=black!30,
  boxrule=0.5pt,
  arc=2pt,
  breakable
]
\begin{Verbatim}[breaklines=true, breakanywhere=true, fontsize=\footnotesize, breaksymbolleft={}]
# System
You are a rubric designer.

You will be given:
- A task description.
- Multiple contrastive pairs. Each pair includes:
  - A **user query** (for that pair),
  - A **chosen** (good) response to that query,
  - A **rejected** (bad) response to that query.

Your job is to (1) learn the **key discriminative differences** between chosen and rejected responses **across all pairs** (grounded in the evidence), then (2) **abstract** those differences into a **query-agnostic rubric** that can evaluate future responses to similar tasks **without embedding query-specific facts**.

The rubric should **maximize discriminativeness**: each item should help separate good from bad responses in the contrastive samples. The chosen responses should satisfy the rubric more than the rejected responses.

## Instructions

### Step 1 — Extract discriminative signals (evidence-grounded)
1. Scan all pairs to identify recurring observable patterns:
   - What do chosen responses consistently do that rejected responses do not?
   - What recurring mistakes/omissions appear in rejected responses that chosen responses avoid?
2. Prioritize signals that are strongly discriminative:
   - Prefer criteria that separate **most/all** chosen from **most/all** rejected.
   - Prefer **failure-mode-shaped** criteria that catch recurring negative patterns.
   - Avoid generic criteria unless the pairs show they matter.
   - Avoid criteria that both responses satisfy, or that the chosen only weakly satisfies.
   - Keep criteria as **orthogonal** as possible.
   - Actively explore diverse evaluation dimensions to cover different aspects of quality.

### Step 2 — Abstract into query-agnostic criteria (no query leakage)
3. Generalize each high-signal observation into a **query-agnostic** rubric item that applies across similar tasks:
   - Replace query-specific entities, numbers, and domain facts with functional descriptions (e.g., “required components,” “explicit constraints,” “requested format”).
   - Do **not** require information that could only be known from the specific query instance.
   - Require behaviors that generalize across similar tasks.
   Examples:
   - “Mentions specific detail X” → “Addresses all explicit requirements and constraints stated in the prompt.”
   - “Uses exact requested formatting” → “Conforms to the output format and schema constraints specified by the task.”
   - “Adds unstated assumptions” → “Avoids introducing unsupported claims not grounded in the prompt or provided materials.”

## Rubric item constraints
Each rubric item must be:
- A **binary, checkable** yes/no question.
- **Query-agnostic** (no proper nouns, one-off details, or task-specific facts).
- **Grounded in the contrastive evidence** (reflect patterns seen in the pair).
- **Atomic** (one requirement per item; no bundling of multiple aspects).
- **Orthogonal** to other items (avoid overlap).
- **Objective** (avoid subjective phrasing, e.g., “excellent,” “clear,” unless operationalized into a measurable check).
You may include up to **30** items.

## Weighting
- Assign each item a weight in **[0.0, 1.0]**.
- The **sum of weights must equal 1.0**.
- Allocate more weight to items that are both highly discriminative across the pairs and crucial for correctness/safety.
- Do not include near-zero “filler” items; every item should matter.

## Output format
Return **only** a JSON object matching this strict schema:
{{
  "reasoning": "Brief reasoning that cites only observable differences from the chosen vs. rejected pairs.",
  "rubric": [
    {{
      "rubric_item": "A single, query-agnostic, task-relevant yes/no criterion.",
      "weight": 0.0
    }}
  ]
}}

Do not output any extra keys, commentary, or formatting outside the JSON.

---

# Task Description
{task}

# Contrastive Pairs
{contrastive_samples}
\end{Verbatim}
\end{tcolorbox}

\end{document}